\documentclass[runningheads]{llncs}
\usepackage{graphicx}
\begin{document}
\title{Deep Learning based Early Detection and Grading of Diabetic Retinopathy Using Retinal Fundus Images}
\titlerunning{Deep Learning based Early Detection and Grading of Diabetic Retinopathy}
%
\author{Sheikh Muhammad Saiful Islam \inst{1} \and
Md Mahedi Hasan \inst{2} \orcidID{0000-0002-6337-3571} \and
Sohaib Abdullah \inst{3}}
\authorrunning{S.M.S. Islam et al.}
%
\institute{Department of Pharmacy, Manarat International University \and
Institute of Information and Communication Technology, Bangladesh University of Engineering and Technology \and
Department of Computer Science and Engineering, Manarat International University \\
\email{saifulislam@manarat.ac.bd, mahedi0803@gmail.com, sohaib@manarat.ac.bd}}

%
\maketitle              

\begin{abstract}
Diabetic Retinopathy (DR) is a constantly deteriorating disease, being one of the leading causes of vision impairment and blindness. Subtle distinction among different grades and existence of many significant small features make the task of recognition very challenging. In addition, the present approach of retinopathy detection is a very laborious and time-intensive task, which heavily relies on the skill of a physician. Automated detection of diabetic retinopathy is essential to tackle these problems. Early-stage detection of diabetic retinopathy is also very important for diagnosis, which can prevent blindness with proper treatment. In this paper, we developed a novel deep convolutional neural network, which performs the early-stage detection by identifying all microaneurysms (MAs), the first signs of DR, along with correctly assigning labels to retinal fundus images which are graded into five categories. We have tested our network on the largest publicly available Kaggle diabetic retinopathy dataset, and achieved $0.851$ quadratic weighted kappa score and $0.844$ AUC score, which achieves the state-of-the-art performance on severity grading. In the early-stage detection, we have achieved a sensitivity of $98\%$ and specificity of above $94\%$, which demonstrates the effectiveness of our proposed method. Our proposed architecture is at the same time very simple and efficient with respect to computational time and space are concerned.

\end{abstract}

\section{Introduction}
\label{introduction}
Diabetic retinopathy, a chronic, progressive eye disease, has turned out to be one of the most common causes of vision impairment and blindness especially for working ages in the world today~\cite{Congdon}. It results from prolonged diabetes. Blood vessels in the light-sensitive tissue (i.e. retina) are mainly affected in diabetic retinopathy. The non-proliferative diabetic retinopathy (NPDR) occurs when the blood vessels leak the blood in the retina. The Proliferative DR (PDR), which causes blindness in the patient, is the next stage to NPDR.

\begin{figure}
	\centerline{\includegraphics[width=90mm]{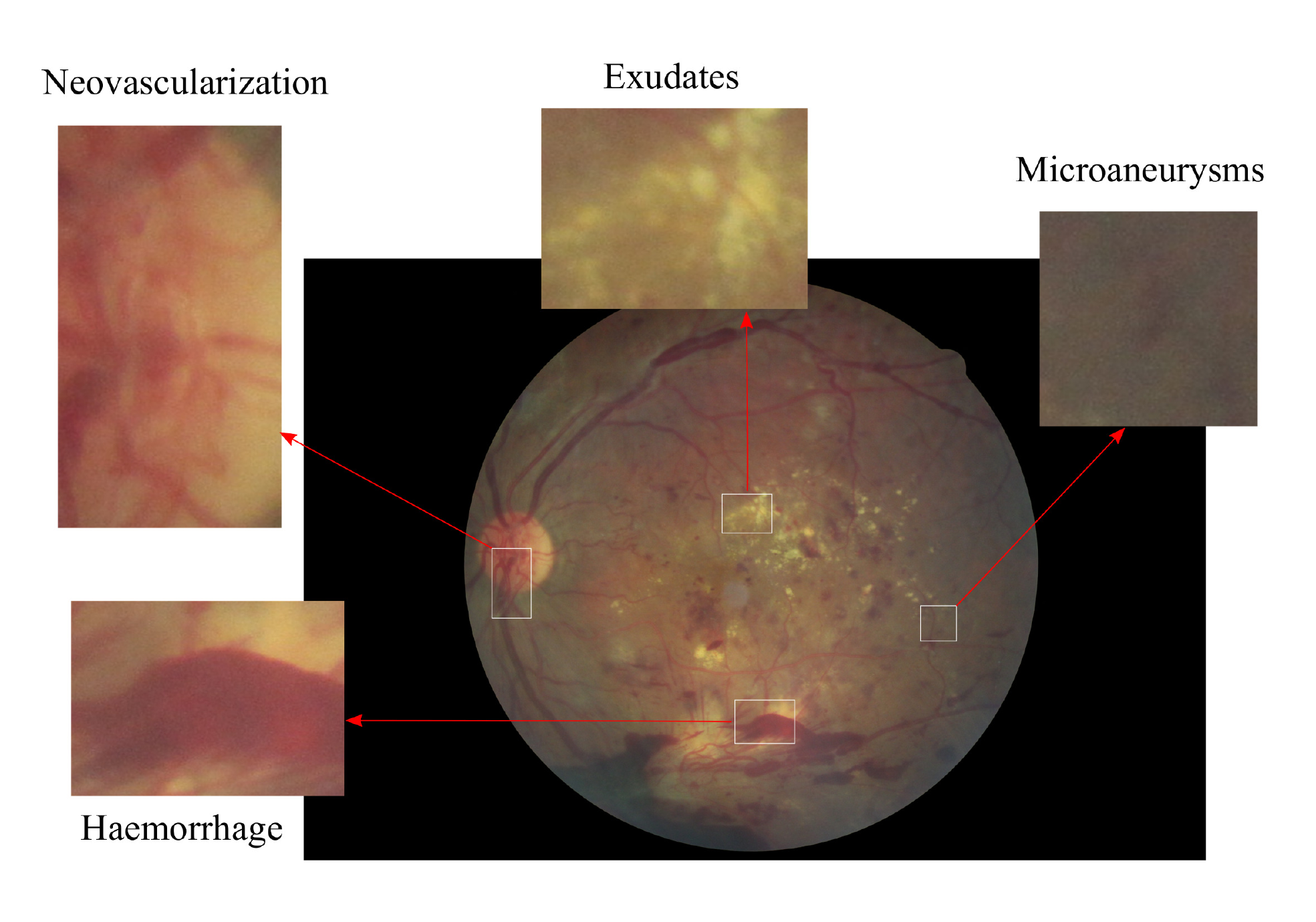}}
	\caption{Example eye image of the proliferative diabetic retinopathy. Additional new blood vessels will begin to grow on the surface of the retina. Due to their abnormal and fragile nature, retinal hemorrhages and ruptured blood vessels are created in this stage which will lead to permanent vision loss.}
	\label{fig_ret_features} 
	
\end{figure}

The progress of DR can be categorized into four stages: mild, moderate, severe nonproliferative diabetic retinopathy, and the advanced stages of proliferative diabetic retinopathy. In mild NPDR, small areas in the blood vessels of the retina, called microaneurysms, swell like a balloon. In moderate NPDR, multiple microaneurysms, hemorrhages, and venous beading occur, causing the patients to lose their ability to transport blood to the retina. The third stage,  called severe NPDR, results from the presence of new blood vessels, which is caused by the secretion of growth factor. The worst stage of DR is the proliferative diabetic retinopathy, as illustrated in Fig.~\ref{fig_ret_features} in which fragile new blood vessels and scar tissue form on the surface of the retina, increasing the likelihood of blood leaking, leading to permanent vision loss.

At present, retinopathy detection system is accomplished by involving a well-trained physician manually detecting vascular abnormalities and structural changes of retina in the retinal fundus images, which are then taken by dilating the retina using vasodilating agent. Due to the manual nature of DR screening methods, however, highly inconsistent results are found from different readers, so automated diabetic retinopathy diagnosis techniques are essential for solving these problems.

Although DR can damage retina without showing any indication at the preliminary stage~\cite{Melville}, successful early-stage detection of DR can minimize the risk of progression to more advanced stages of DR. The diagnosis is particularly difficult for early-stage detection because the process relies on discerning the presence of microaneurysms, retinal hemorrhages, among other features on the retinal fundus images. Furthermore, accurate detection and determination of the stages of DR can greatly improve the intervention, which ultimately reduces the risk of permanent vision loss.

Earlier solutions of automated diabetic retinopathy detection system were based on hand-crafted feature extraction and standard machine learning algorithm for prediction~\cite{Silberman}. These approaches were greatly suffer due to the hand-crafted nature of DR features extraction since feature extraction in color fundus images are more challenging compared to the traditional images for object detection task. Moreover, these hand-crafted features are highly sensitive to the quality of the fundus images, focus angle, presence of artifacts, and noise. Thus, these limitations in traditional hand-crafted features make it important to develop an effective feature extraction algorithm to effectively analyze the subtle features related to the DR detection task.

In recent times, most of the problems of computer vision have been solved with greater accuracy with the help modern deep learning algorithms, Convolutional Neural Networks (CNNs) being an example. CNNs have been proven to be revolutionary in different fields of computer vision such as object detection and tracking, image and medical disease classification and localization, pedestrian detection, action recognition, etc. The key attribute of the CNN is that it extracted features in task dependent and automated way. So, in this paper, we present an efficient CNN architecture for DR detection in large-scale database. Our proposed network was designed with a multi-layer CNN architecture followed by two fully connected layer and an output layer. Our network outperforms other state-of-the-art network in early-stage detection and achieves state-of-the-art performance in severity grading of diabetic retinopathy detection.

The rest of the paper is organized as follows: related work is presented in the section~\ref{related_work}, followed by the proposed method in the section~\ref{proposed_method}, while experimental set-up and results are discussed in the section~\ref{exp_results}. Finally, we draw our conclusion in the section~\ref{conclusion}.

\section{Related Work } 
\label{related_work}

The earlier works on automatic diabetic retinopathy detection were based on designing hand-crafted feature detectors to measure the blood vessels and optic disc, and on counting the presence of abnormalities such as microaneurysms, red lesions, hemorrhages, and hard exudates, etc. The detection was performed using these extracted features by employing various machine learning methods like support vector machines (SVM) and  k-nearest neighbor (kNN)~\cite{Silberman,Sopharak1}. In~\cite{Acharya}, Acharya et al. used features of blood vessel area, microaneurysms, exudes, and hemorrhages with an SVM, achieving an accuracy of $86\%$, specificity of $ 86\% $, and sensitivity of $ 82\% $. Roychowdhury et al.~\cite{Roychowdhury} developed a two-step hierarchical classification approach, where the non-lesions or false positives were rejected in the first step. For lesion classification in the second step, they used classifiers such as the Gaussian mixture model (GMM), kNN, and support vector machine (SVM). They achieved sensitivity of $ 100\% $, specificity of $ 53.16\% $, and AUC $ 0.904 $. However, these types of approaches have the disadvantage of utilizing limited number of features.

Deep learning based algorithms have become popular in the last few years. For example, standard ImageNet architectures were used in~\cite{Pratta,swang}. Furthermore, Kaggle~\cite{kaggle} has recently launched a DR detection competition, where all the top-ranked solutions were implemented employing CNN as the key algorithm. Pratta et al.~\cite{Pratta} developed a CNN based model, which surpassed human experts in classifying advanced stages of DR. In~\cite{antal}, CNN based method was employed to detect microaneurysms a DR stage grading. Ensemble of CNN was employed to simultaneously detect DR and macular edema by Kori et al.~\cite{kori}. They employed a variant of ResNet~\cite{kaiming} and densely connected networks~\cite{Gao1}. To make the model prediction more interpretable, a visual map was generated by Torre et al.~\cite{Torre} using CNN model, which can be used to detect lesion in the tested retinal fundus images. A similar approach was used in~\cite{Wang1} along with generation of regression activation map (RAM). 

Some researches focused on exploring breakdown of classification task into subproblem prediction tasks. For example, Yang et al.~\cite{Yang1} employed a two-stage deep convolutional neural network based methodology, where exudates, microaeurysms,and haemorrhage were first detected by local network and subsequent severity grading was performed by global network. By introducing unbalanced weight map to emphasize leison detection, they achieved AUC of $0.9590$. Authors of~\cite{Bravo1} implemented an architecture like VGG-16~\cite{Simonyan} and Inception-4~\cite{Szegedy} network for DR classification. 

Some recent works~\cite{Butterworth} in diabetic retinopathy have leveraged mean squared error objective function to convert the classification task into a regression task. Here ensemble of classical machine learning algorithms, like naive bayes classifier, SVM, as well as ImageNet state-of-the-art networks, with mean squared error objective function applied to tackle DR detection problem with accuracy, Kappa, and F-score of $0.736$, $0.676$, and $0.417$, respectively.

\section{Proposed Method}
\label{proposed_method}

\subsection{Data Preprocessing}
\begin{figure}
	\centerline{\includegraphics[width = 90mm]{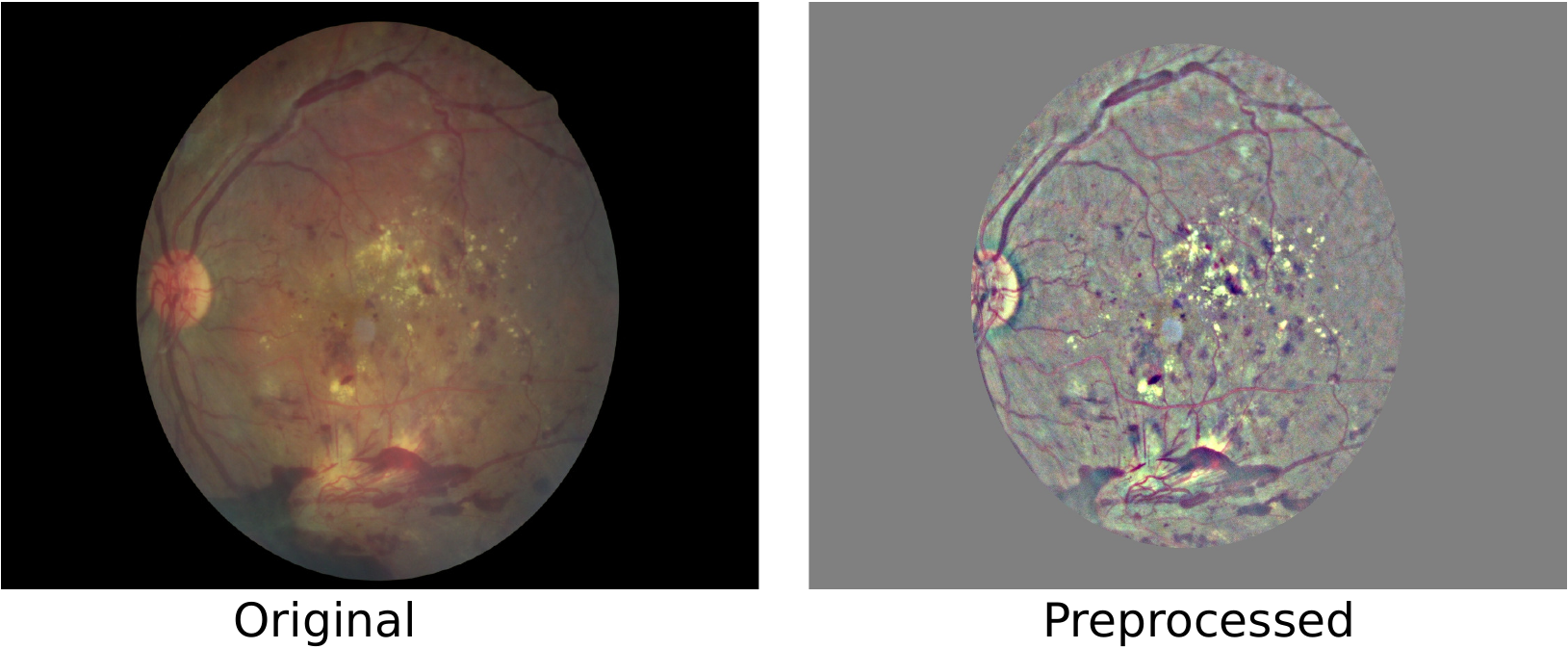}}
	\caption{Demonstration of the retinal fundus image before and after preprocessing.}
	\label{fig_preprocess}	
\end{figure}
There exists a lot of exposure and lighting variation over the original fundus images, so we took several preprocessing steps as suggested by Graham ~\cite{Graham} to  standardize the image condition. First, we re-scaled the images to get the same radius and subtracted the local average color. Thereafter, the local average color of the images were mapped to $50\%$ gray. We also clipped the images to $90\%$ size to remove the boundary effects. The sample of the resulted preprocessed images, along with the original images, are illustrated in Fig.~\ref{fig_preprocess}.

\subsection{Data Augmentation}
\begin{figure}
	\centerline{\includegraphics[width = 120mm]{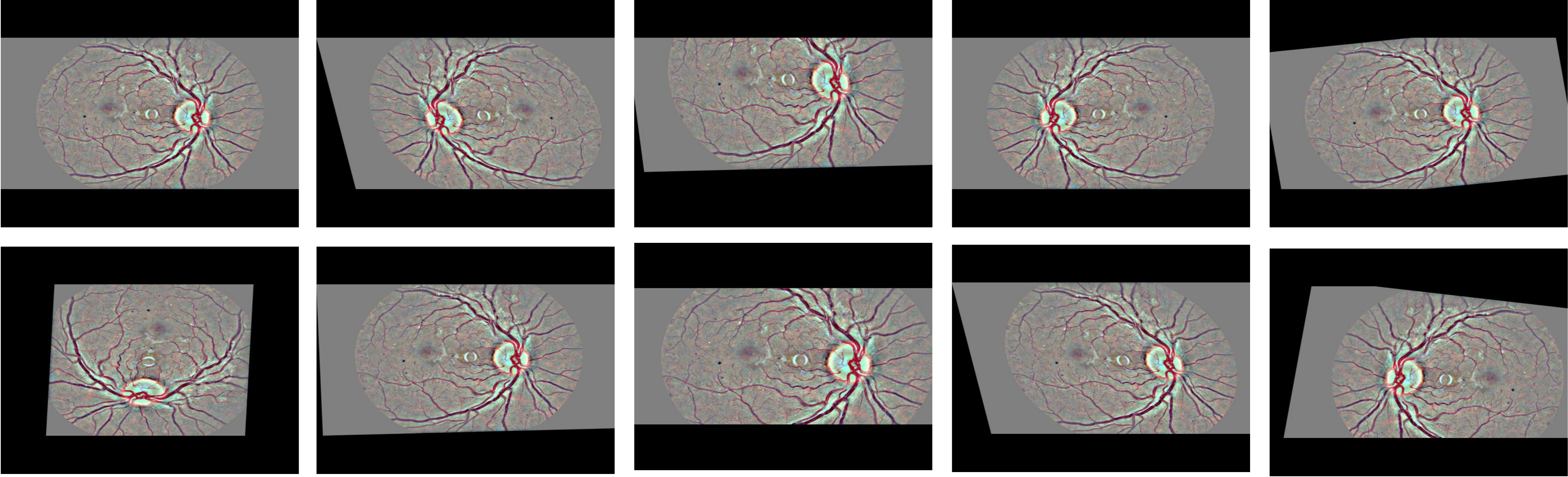}}
	\caption{Examples of some augmentation operation performed on a preprocessed retinal images. After the operation, each augmented image is resized maintaining the aspect ratio.}
	\label{fig_aug}
\end{figure}
The performance of deep neural network is strongly correlated with the size of available training data. Although Kaggle EyePACS dataset is largest for retinopathy detection consisting of around $88,702$ images, We are to use a very small fraction of it containing images for disease severity grading task with imbalanced classes, requiring us to heavily augment our training data to obtain a model which is stable and not overfitted. The major data augmentation operations that we performed are listed below.

\begin{table}
	\begin{center}
		\caption{Statistics of the augmentation operations performed on the training set of Kaggle EyePACS dataset. Due to highly imbalanced nature of the dataset, different grades were augmented differently.}
		\label{tbl_dataset}
		\begin{tabular}{cccc|cc} 
			\textbf{Grade} & \quad \textbf{Raw} & \textbf{Training} &\quad \textbf{Validation} &\quad \textbf{Operations} & \quad \textbf{Total}   \\[0.4mm]
			
			\hline
			\rule{0pt}{3ex}
			Normal & \quad 25810 & 25610 & \quad 200  & \quad 0 & 25810  \\[0.4mm]
			
			\hline
			\rule{0pt}{3ex}
			Mild NPDR & \quad 2443 & 2243 & \quad 200 & \quad 11 & 26916   \\[0.4mm]
			
			\hline
			\rule{0pt}{3ex}
			Moderate NPDR & \quad 5292 & 5092 & \quad 200 & \quad 4 & 25460     \\[0.4mm]
			
			\hline
			\rule{0pt}{3ex}
			Severe NPDR & \quad 873 & 673 & \quad 200 & \quad 27 & 18844      \\[0.4mm]				
			
			\hline
			\rule{0pt}{3ex}
			Proliferative DR & \quad 708 & 508 & \quad 200 & \quad 35 & 18288      \\[0.4mm]
			
			\hline
			\rule{0pt}{3ex}
			\textbf{Total}& \quad $35,126$ & \quad $34,126$ & \quad $1,000$ &  & $115,318$   \\[0.4mm]
						
		\end{tabular}
	\end{center}
\end{table}
\begin{itemize}
	\item Rotation :– Images were randomly rotated between $0^\circ$ to $360^\circ$  
	\item Shearing :-  Randomly sheared with angle between −20$^\circ$ and 20$0^\circ$
	\item Flip :– Images were both horizontally and vertically flipped
	\item Zoom :– Images were randomly stretched between (1/1.3 , 1.3)
	\item Crop :- Images were randomly cropped to $85-95\%$ of the original size
	\item Krizhevsky augmentation :– Images were augmented by Krizhevsky color augmentation technique~\cite{Krizhevsky}.
	\item Translation :– Images were randomly shifted between $-25$ and $25$ pixel
\end{itemize}

Also, we scaled and centered each image channel (RGB) to get zero mean and unit variance over the dataset. Fig.~\ref{fig_aug} shows some post-augmentation example images. We eventually obtained up to 35 augmentation version of each training image. The output image sizes of our data augmentation pipeline are $512$ x $512$ and $448$ x $448$.

\subsection{Network Architecture}
\begin{table}
	\begin{center}
		\caption{The proposed network architecture of our early-stage detection and severity grading model. The depth of the network is 18 layer while the kernel size of the convolutional filter is $4$ x $4$. Max-pooling of $3$ x $3$ is used to downsample the activation map. After convolution layer, two fully connected layers of 1024 neuron followed by a single neuron output layer are added considering the early-stage detection and severity grading task as a regression problem.}
		
		\label{tbl_network}
		\begin{tabular}{cccc} 
			\textbf{Layer Type} & \quad\textbf{Kernel Size \& Number} & \quad\textbf{Stride} & \quad\textbf{Output Shape}  \\[0.4mm]
			
			\hline
			\rule{0pt}{3ex}
			input &  ..  & .. & $(512, 512, 3)$ \\
			
			\hline
			\rule{0pt}{3ex}
			convolution & 4 x 4 x 32 & 2 & $(256, 256, 32)$ \\
			
			\hline
			\rule{0pt}{3ex}
			convolution & 4 x 4 x 32 & 1 & $(255, 255, 32)$ \\
			
			\hline
			\rule{0pt}{3ex}
			max-pooling & 3 x 3 & 2 & $(127, 127, 32)$ \\				
			
			\hline
			\rule{0pt}{3ex}
			convolution & 4 x 4 x 64 & 2 & $(62, 62, 64)$ \\
			
			\hline
			\rule{0pt}{3ex}
			convolution & 4 x 4 x 64 & 1 & $(63, 63, 64)$ \\
			
			\hline
			\rule{0pt}{3ex}
			max-pooling & 3 x 3 & 2 & $(31, 31, 64)$ \\	
			
			\hline
			\rule{0pt}{3ex}
			convolution & 4 x 4 x 128 & 1 & $(32, 32, 128)$ \\
			
			\hline
			\rule{0pt}{3ex}
			convolution & 4 x 4 x 128 & 1 & $(33, 33, 128)$ \\
			
			\hline
			\rule{0pt}{3ex}
			max-pooling & 3 x 3 & 2 & $(16, 16, 128)$ \\	
			
			\hline
			\rule{0pt}{3ex}
			convolution & 4 x 4 x 256 & 1 & $(17, 17, 256)$ \\
			
			\hline
			\rule{0pt}{3ex}
			max-pooling & 3 x 3 & 2 & $(8, 8, 256)$ \\
			
			\hline
			\rule{0pt}{3ex}
			convolution & 4 x 4 x 384 & 1 & $(9, 9, 384)$ \\
			
			\hline
			\rule{0pt}{3ex}
			max-pooling & 3 x 3 & 2 & $(4, 4, 384)$ \\
			
			\hline
			\rule{0pt}{3ex}
			convolution & 4 x 4 x 512 & 1 & $(5, 5, 512)$ \\
			
			\hline
			\rule{0pt}{3ex}
			max-pooling & 3 x 3 & 2 & $(2, 2, 512)$ \\
			
			\hline
			\rule{0pt}{3ex}
			fully connected & 1024 & .. & $(1024)$ \\
			\hline
			\rule{0pt}{3ex}
			fully connected & 1024 & .. & $(1024)$ \\
			\hline
			\rule{0pt}{3ex}
			fully connected & 1 & .. & $(1)$ \\
			\hline
			
		\end{tabular}
	\end{center}
\end{table}

Table~\ref{tbl_network} illustrates the network architecture of our proposed DR detection method. The input layer of the network is $512$ x $512$. We tried several kernel filters of sizes 3 x 3, 4 x 4, 5 x 5 and found the best result in kernel of size 4 x 4. Hence, all the convolutional layers of our network have the kernel size of 4 x 4 with united bias and the stride of 1 except the first and third convolutional layer which have stride of 2. 

LeakyReLU~\cite{Maas} was used in all convolutional layer as the activation function for nonlinearity. All the Max-Pooling layers used have same kernel size of 3 x 3. The final extracted local features were flattened before passing through fully connected layers. There are two fully connected layers, each having 1024 neurons. Dropout of $0.5$ was added after all but the last fully connected layers to reduce overfitting. Since it is much worse to misclassify severe NPDR or PDR as normal eye than as moderate retinopathy, we considered this multi-class classification as a regression problem and so an output layer of one neuron was added. We took \textit{mean squared error} as our objective function. Also, we clipped the loss function value between 0 and 4 since our class ranges between these values.

\begin{table}
	\begin{center}
		\caption{Network architecture of features blending network}
		\label{tbl_blend_network}
		\begin{tabular}{cccc} 
			\textbf{Layer Type} & \quad\textbf{Output Shape}  \\[0.4mm]
			
			\hline
			\rule{0pt}{3ex}
			 Input layer & $(4096)$ \\
			 
			\hline
			\rule{0pt}{3ex}
			 Fully connected & $(32)$ \\
			\hline
			\rule{0pt}{3ex}
			 Maxout network & $(16)$ \\
			 
			 \hline
			 \rule{0pt}{3ex}
			 Fully connected & $(32)$ \\
			 \hline
			 \rule{0pt}{3ex}
			 Maxout network & $(16)$ \\
			 
			 \hline
			 \rule{0pt}{3ex}
			 Fully connected & $(1)$ 
		\end{tabular}
	\end{center}
\end{table}

In diabetic retinopathy experiment, blending the features for both eyes of a patient usually leads to a significant 
improvement in performance~\cite{Antony}. Thus, we blended our features according to the state-of-the-art blending method~\cite{Antony}, where the output of our last max-pooling layer was also used as input features to the blending network, as shown in Table~\ref{tbl_blend_network}. To improve the feature quality, feature extraction was repeated as many as 40 times with different augmentations per image. The mean and standard deviation of each feature was used as input to our blending network.

\subsection{Training}
Our eighteen-layer-deep proposed network has more than 8.9 million parameters which were randomly initialized using 
orthogonal weight initialization. The network was trained with SGD optimization function with 0.90 Nesterov momentum with a fixed schedule over $300$ epochs with data augmentation at each step. L2 regularization with a factor of 0.0005 was applied to all weighted layers. We tested several learning rates but found $10^{-4}$ to be the best initial learning rate. The learning rate constantly decreased as the number of training epochs progressed and ended up having a learning rate of $10^{-6}$. The summary of the settings of our training hyperparameters are given in Table~\ref{tbl_hyper_sg}.

\begin{table}
	\begin{center}
		\caption{Hyperparameters setting of our proposed network architecture}
		\label{tbl_hyper_sg}
		\begin{tabular}{cc} 
			\textbf{ Hyperparameter} & \textbf{Value} \\
			\hline
			\hline \rule{0pt}{3ex}
			Objective function & Mean Squared Error (MSE) \\[0.4mm]

			\hline \rule{0pt}{3ex}
			Optimizer & SGD \\[0.4mm]
			
			\hline \rule{0pt}{3ex}
			Momentum & 0.9\\[0.4mm]
			
			\hline \rule{0pt}{3ex}
			Multiple learning rates & {
				\begin{tabular}{c}
					${10}^{-4}$ (for 80 epochs) \\ [0.4mm] \hline \rule{0pt}{3ex}
					${10}^{-5}$ (for next 70 epochs) \\ [0.4mm]\hline \rule{0pt}{3ex}
					${5}$x${10}^{-5}$ (for next 40 epochs) \\ [0.4mm]\hline \rule{0pt}{3ex}
					${10}^{-6}$ (for next 110 epochs) 
				\end{tabular}
			} \\ [0.4mm]
			
			\hline \rule{0pt}{3ex}
			Batch size & 16 \\[0.4mm]
			
			\hline \rule{0pt}{3ex}
			Epoch & 300 \\[0.4mm]
			
			\hline
		\end{tabular}
	\end{center}
\end{table}

The blending network was trained with Adam~\cite{Kingma} optimization algorithm with a fixed schedule over $100$ epochs. ReLU activation function was used after each fully connected layer and an L2 regularization with a factor of 0.001 was applied to every layer. The batch size used was 32 considering \textit{mean squared error} as our objective function.

\section{Experimental Results}
\label{exp_results}

\subsection{Dataset}
There are several publicly available databases of fundus images for DR including DIARETDB0~\cite{Kauppi}, DIARETDB1~\cite{Kamarainen}, Kaggle EyePACS~\cite{kaggle}, and Messidor~\cite{Decencière} Databases. DIARETDB0 dataset contains 130 color fundus images, out of which 20 are normal and 110 are affected by DR. On the other hand, DIARETDB1 contains a total 89 color fundus images, out of which 84 images contain sign of microaneurysms and five images are normal. The Messidor database contains 1200 retinal fundus images.

\begin{figure}
	\centerline{\includegraphics[width=120mm]{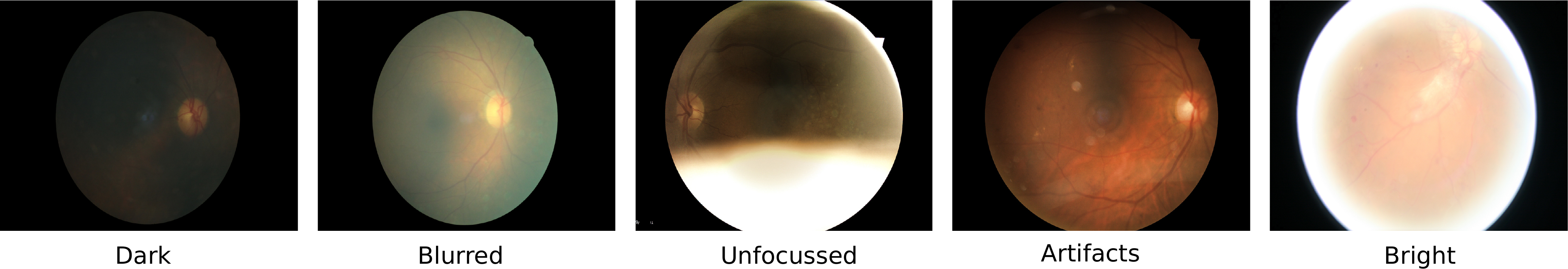}}
	\caption{Example of some poor quality retinal images of Kaggle EyePACS dataset including too bright, dark, and blurred images which make it difficult for the learning algorithm to accurately classify the DR grades.}
	\label{fig_deficient_images}
\end{figure}

In this work, we used EyePACS, the largest publicly available dataset for diabetic retinopathy, from Kaggle Diabetic Retinopathy Detection competition~\cite{kaggle} sponsored by the California Healthcare Foundation, which contains a total of $88,702$ high-resolution retinal fundus images, captured under a variety of imaging conditions. The dataset also contains artifacts, out of focus, too bright, and too dark images as illustrated in 
Fig.~\ref{fig_deficient_images}.

Fundus images are categorized into five grades $(0, 1, 2, 3, 4)$ according to the severity of DR which are named sequentially as healthy or normal image, mild non-proliferative DR, moderate non-proliferative DR, severe non-proliferative DR, and proliferative DR. The dataset is highly imbalanced due to the presence of around $75\%$  grade 0 (no DR) images. Table~\ref{tbl_dataset} shows overview of the distribution among different grades in the training set images of the dataset. We split the dataset with $35,126$ images in training set and $53,576$ images in test set as suggested by the Kaggle competition~\cite{kaggle} dataset settings. We also used $96\%$ of images of each class in training set for training and $4\%$ for validation. 

\begin{table}
	\begin{center}
		\caption{Grade distribution in the training set of Kaggle EyePACS dataset}
		\label{tbl_dataset}
		\begin{tabular}{cccc} 
			\textbf{DR Grade} & \quad\textbf{Grade Name} & \quad\textbf{Total Images} & \quad\textbf{Percentage }  \\[0.4mm]
			
			\hline
			\rule{0pt}{3ex}
			0 &  Normal & 25810 & 73.84\% \\[0.4mm]
			
			\hline
			\rule{0pt}{3ex}
			1 & Mild NPDR & 2443 & 6.96\% \\[0.4mm]
			
			\hline
			\rule{0pt}{3ex}
			2 & Moderate NPDR & 5292 & 15.07\% \\[0.4mm]
			
			\hline
			\rule{0pt}{3ex}
			3 & Severe NPDR & 873 & 2.43\% \\[0.4mm]				
			
			\hline
			\rule{0pt}{3ex}
			4 & Proliferative DR & 708 & 2.01\% \\[0.4mm]			
		\end{tabular}
	\end{center}
\end{table}

\subsection{Performance Evaluation on Early-Stage Detection}
In this work, we performed two binary classification for early-stage detection experiment: sick (grades $1,2,3,4$) vs healthy (grade 0), and low (grades $0,1$) vs high (grades $2,3,4$). We considered both of these sub-problems equally important for early-stage detection.

\begin{table}
	\begin{center}
		\caption{Performance evaluation of our proposed method on DR early-stage detection. Considering early-stage detection as a binary classification problem, our proposed method achieved astonishing $98\%$ sensitivity and $94\%$ in low-high DR detection.}
		\label{tbl_performance_ed}
		
		\begin{tabular}{ccc} 
			\textbf{Classification Problem} & \quad\textbf{Sensitivity} & \quad\textbf{Specificity} 
			\\[0.4mm] \hline
			
			\rule{0pt}{3ex}
	    	Healthy (0) vs Sick (1,2,3,4)  & 94.5\% & 90.2\% \\[0.4mm] \hline
	    	
	    	\rule{0pt}{3ex}
	    	Low (0,1) vs High (2,3,4) & 98\%   & 94\%
		\end{tabular}
	\end{center}
\end{table}

We calculated sensitivity and specificity metric for both of these binary classification problems and found out higher performance in low-high DR classification than in healthy-sick classification. This is because retinal features of grade $(0, 1)$ are similar to grade $(2, 3, 4)$

\subsection{Performance Evaluation on Severity Grading}
The \textit{quadratic weighted kappa}, the state-of-the-art performance metric for multi-class classification and suggested evaluation metric for DR~\cite{kaggle}, is adopted as the performance metric of our severity grading prediction. We took thresholds $(0.5, 1.5, 2.5, 3.5)$ to discretize the predicted regression values and make the class levels into integer for computing the Kappa scores. We achieved 0.851 quadratic weighted kappa on test set of Kaggle dataset after submitting our solution in~\cite{kaggle}.

\begin{table}
	\begin{center}
		\caption{Performance evaluation of our proposed method on severity grading of diabetic retinopathy.}
		\label{tbl_performance_sg}
		\begin{tabular}{ccc} 
			
			\textbf{Metrics} &\qquad\textbf{Mean squared error} \\[0.4mm]
			
			\hline
			\rule{0pt}{3ex}
			 Quadratic Weighted Kappa &\quad 0.851  \\[0.4mm]
			
			\hline
			\rule{0pt}{3ex}
			Area Under the ROC Curve &\quad 0.844 \\[0.4mm]
			
			\hline
			\rule{0pt}{3ex}
			F-Score & \quad 0.743  \\[0.4mm]
		\end{tabular}
	\end{center}
\end{table}

We also calculated Area Under the ROC Curve (AUROC) and F-Score of our proposed architecture on the same dataset and achieved score $0.844$ and $0.743$ respectively. 

\section{Conclusion}
\label{conclusion}
In this paper, we have presented a novel CNN-based deep neural network to detect early-stage and severity grades of diabetic retinopathy in retinal fundus images. In our work, we have found that without heavy data augmentation, a high capacity network can easily overfit the training data. Even with data-augmentation, any network can overfit on oversampled classes such as healthy eye (grade 0). Thus, designing a small capacity network with L2 regularization, and dropout has significant importance in retinopathy detection. So, in this work, we have presented a 4 x 4 kernel based CNN network with several preprocessing and augmentation methods to improve the performance of the architecture. Our network achieved 98\% sensitivity and more than 94\% specificity in early-stage detection and a kappa score of more than 0.85 in severity grading on the challenging Kaggle EyePACS dataset. The experimental results have demonstrated the effectiveness of our proposed algorithm to be good enough to be employed in clinical applications.

%
%
%
%

\end{document}